# Multitask Kernel-based Learning with Logic Constraints

**Michelangelo Diligenti, Marco Gori, Marco Maggini, Leonardo Rigutini** [1]

**Abstract.** This paper presents a general framework to integrate prior knowledge in the form of logic constraints among a set of task functions into kernel machines. The logic propositions provide a partial representation of the environment, in which the learner operates, that is exploited by the learning algorithm together with the information available in the supervised examples. In particular, we consider a multi-task learning scheme, where multiple unary predicates on the feature space are to be learned by kernel machines and a higher level abstract representation consists of logic clauses on these predicates, known to hold for any input. A general approach is presented to convert the logic clauses into a continuous implementation, that processes the outputs computed by the kernel-based predicates. The learning task is formulated as a primal optimization problem of a loss function that combines a term measuring the fitting of the supervised examples, a regularization term, and a penalty term that enforces the constraints on both supervised and unsupervised examples. The proposed semi-supervised learning framework is particularly suited for learning in high dimensionality feature spaces, where the supervised training examples tend to be sparse and generalization difficult. Unlike for standard kernel machines, the cost function to optimize is not generally guaranteed to be convex. However, the experimental results show that it is still possible to find good solutions using a two stage learning schema, in which first the supervised examples are learned until convergence and then the logic constraints are forced. Some promising experimental results on artificial multi-task learning tasks are reported, showing how the classification accuracy can be effectively improved by exploiting the a priori rules and the unsupervised examples.

## 1 Introduction

Learning machines can significantly benefit from incorporating prior knowledge about the environment into a learning schema based on a collection of supervised and unsupervised examples. Remarkable approaches to provide a unified treatment of logic and learning consist of integrating logic and probabilistic calculus, which gave rise to the field of probabilistic inductive logic programming. In particular, [7] proposes to use support vector machines with a kernel that is an inner product in the feature space spanned by a given set of first-order hypothesized clauses. Frasconi et al. [4] provide a comprehensive view on statistical learning in the inductive logic programming setting based on kernel machines, in which the background knowledge is injected into the learning process by encoding it into the kernel function.

This paper proposes a novel approach to incorporate an abstract and partial representation of the environment in which the learner operates, in the form of a set of logic clauses, which are expected to impose constraints on the development of a set of functions that are to be inferred from examples. We rely on a multi-task learning scheme [2], where each task corresponds to a unary predicate defined on the feature space, and the domain knowledge is represented via a set of FOL clauses over these task predicates. The kernel machine mathematical apparatus allows us to approach the problem as primal optimization of a cost function composed of the loss on the supervised examples, the regularization term, and a penalty term that forces the constraints coupling the learning tasks for the different predicates. Well established results can be used to convert the logic clauses in a continuous form, yielding a constrained multi-task learning problem. Once the constraint satisfaction is relaxed to hold only on the supervised and unsupervised examples, a representation theorem holds dictating that the optimal solution of the problem is a kernel expansion over these examples.

Unlike for classic kernel machines, the error function is not guaranteed to be convex, which clearly denotes the emergence of additional complexity. Following inspirations coming from the principles of cognitive development stages, that have been the subject of an in-depth analysis in children by J. Piaget, the experimental results show evidence that an ad-hoc stage-based learning as sketched in [5] allows the discovery of good solutions to complex learning tasks. This also suggests the importance of devising appropriate teaching plans like the one exploited in curriculum learning [1]. In our setting, pure learning from the supervised examples is carried out until convergence and, in a second stage, learning continues by forcing the logic clauses. Because of the coherence of supervised examples and logic clauses, the first stage facilitates significantly the optimization of the penalty term, since classic gradient descent heuristics are likely to start closer to the basin of attraction of the global minimum with respect to a random start. The experimental results compare the constraint based approach against plain kernel machines on artificial learning tasks, showing that the proposed semi-supervised learning framework is particularly suited for learning in high dimensionality input spaces, where the supervised training examples tend to be sparse and generalization difficult.

[1] University of Siena, Italy, email: {diligmic,marco,maggini,rigutini}@dii.unisi.it

## 2 Learning with constraints

We consider a multi-task learning problem in which the a set of functions $\{f_k : \mathcal{X} \to I\!R, \quad k = 1, \ldots, T\}$ must be inferred from examples, where $\mathcal{X}$ is a set of objects. For the sake of simplicity, we will consider the case where each task processes the same feature representation $\mathbf{x} = \mathcal{F}(X) \in F \subset I\!R^d$ of an input object $X \in \mathcal{X}$, but the framework can be trivially extended to the case when different feature spaces are exploited for each task. In this paper we restrict our attention to classification, assuming that each function $f_k$ provides an evidence of the input to belong to the corresponding class $k$. We propose to model the prior knowledge on the tasks as a set of constraints on the configurations of the values for $\{f_k(\mathbf{x})\}$, that are implemented by functions $\phi_h \ : \ I\!R^T \to I\!R$:

$$\phi_h(f_1(\mathbf{x}), \ldots, f_T(\mathbf{x})) \geq 0 \quad \forall \mathbf{x} \in F, \quad h = 1, \ldots, H \ . \qquad (1)$$

Let us suppose that each function $f_k$ can be represented in an appropriate Reproducing Kernel Hilbert Space (RKHS) $\mathcal{H}_k$. We employ the classical learning formulation where a set of supervised samples, extracted from the unknown distributions $p_{\mathbf{x}y_k}(\mathbf{x}, y_k)$, correlates the input with the target values $y_k$. The supervised examples are organized in the sets $\mathcal{L}_k = \{(\mathbf{x}_k^i, y_k^i)|i = 1, \ldots, \ell_k\}$, where only a partial set of labels over the tasks can be available for any given sample $\mathbf{x}_k^i$. The unsupervised examples are collected in $\mathcal{U} = \{\mathbf{x}^i : i = 1, \ldots, u\}$, while $\mathcal{S}_k^L = \{\mathbf{x}_k^i : (\mathbf{x}_k^i, y_k^i) \in \mathcal{L}_k\}$ collects the sample points in the supervised set for the $k$-th task. The set of the available points is $\mathcal{S} = \bigcup_k \mathcal{S}_k^L \bigcup \mathcal{U}$.

The learning problem is cast in a semi-supervised framework, that aims at optimizing the cost function:

$$E(\mathbf{f}) = R(\mathbf{f}) + N(\mathbf{f}) + V(\mathbf{f}) \qquad (2)$$

where, in addition to the fitting loss $R(\cdot)$ and the regularization term $N(\cdot)$, the term $V(\cdot)$ penalizes the violated constraints. In particular, the error risk associated with $\mathbf{f} = [f_1, \ldots, f_T]'$ is:

$$R(\mathbf{f}) = \sum_{k=1}^{T} \lambda_k^\tau \cdot \frac{1}{|\mathcal{L}_k|} \sum_{(\mathbf{x},y) \in \mathcal{L}_k} L_k^e(f_k(\mathbf{x}), y),$$

where $L_k^e(f_k(\mathbf{x}), y)$ is a loss function that measures the fitting quality of $f_k(\mathbf{x})$ with respect to the target $y$ and $\lambda_k^\tau > 0$. As for the regularization term, we employ simple scalar kernels,

$$N(\mathbf{f}) = \sum_{k=1}^{T} \lambda_k^r \cdot ||f_k||^2_{\mathcal{H}_k},$$

where $\lambda_k^r > 0$. Please note that the framework could be trivially extended to include multi-task kernels that consider the interactions amongst the different tasks [2]. Finally, the penalty term $V(\cdot)$ taking the constraints into account is defined as:

$$V(\mathbf{f}) = \frac{1}{|\mathcal{S}|} \sum_{h=1}^{H} \lambda_h^v \cdot \sum_{\mathbf{x} \in \mathcal{S}} L_h^c(\phi_h(f_1(\mathbf{x}), \ldots, f_T(\mathbf{x}))) \ ,$$

where $\lambda_h^v > 0$ and the penalty loss function $L_h^c(\phi)$ is strictly positive when the constraint is violated. For instance, a natural choice for the constraints penalty term is the hinge-like function $L_h^c(\phi) = \max(0, -\phi)$. Unlike the previous terms, the constraint penalty involves all the functions simultaneously and introduces a correlation among the tasks in the learning process. Interestingly, the optimal solution of equation (2) can be expressed by a kernel expansion as stated in the following Representer Theorem.

**Theorem 1** *Let us consider a multi-task learning problem for which the task functions $f_1, \ldots, f_T$, $f_k : I\!R^n \to I\!R$, $k = 1, \ldots, T$, are assumed to belong to the RKHSs $\mathcal{H}_1, \ldots, \mathcal{H}_T$. Then the optimal solution $[f_1^*, \ldots, f_T^*] = argmin_{f_1 \in \mathcal{H}_1, \ldots, f_T \in \mathcal{H}_T} E([f_1, \ldots, f_T])$ can be expressed as,*

$$f_k^*(\mathbf{x}) = \sum_{\mathbf{x}^i \in \mathcal{S}} w_{k,i}^* K_k(\mathbf{x}^i, \mathbf{x})$$

*where $K_k(\mathbf{x}', \mathbf{x})$ is the kernel associated to the space $\mathcal{H}_k$.*

*Proof:* The proof is a straightforward extension of the representer theorem for plain kernel machines [8]. It is only sufficient to notice that like for the term corresponding to the empirical risk, also the penalty term enforcing the constraints only involves values of $f_k$ sampled in $\mathcal{S}$. □

This representer theorem allows us to optimize (2) in the primal by gradient descent heuristics [3]. The weights of the kernel expansion can be compactly organized in $\mathbf{w}_k = [w_{k,1}, \ldots, w_{k,|\mathcal{S}|}]'$ and, therefore, the optimization of (2) turns out to involve directly $\mathbf{w}_k, k = 1, \ldots, T$. In order to compute the gradient, let us consider the three different terms separately. Let $\mathbf{K}_k = \left[K_k(\mathbf{x}^i, \mathbf{x}^j)\right]_{i,j=1,\ldots,|\mathcal{S}|}$ be the Gram matrix associated to the kernel and consider the vector

$$\mathbf{dL}_k^e = \left[ \left. \frac{\partial L_k^e(f, y)}{\partial f} \right|_{(f_k(\mathbf{x}^j), y^j)} \right]'_{\mathbf{x}^j \in \mathcal{S}},$$

that collects the loss function derivatives computed for all the samples in $\mathcal{S}$. For the unlabeled samples any value can be set, since they are not involved in the computation. In fact, we introduce the diagonal matrix $\mathbf{I}_k^L$, whose $j$-th diagonal element is set to 1 if the $j$-th sample is supervised, i.e. $\mathbf{x}^j \in \mathcal{S}_k^L$, and 0 otherwise. Hence, we have $\nabla_k R(\mathbf{f}) = \frac{\lambda_k^\tau}{|\mathcal{L}_k|} \cdot \mathbf{K}_k \cdot \mathbf{I}_k^L \cdot \mathbf{dL}_k^e$ . Likewise, the gradient of $N(\mathbf{f})$ can be written as $\nabla_k N(\mathbf{f}) = 2 \cdot \lambda_k^r \cdot \mathbf{K}_k \cdot \mathbf{w}_k$ . Finally, if we define

$$\mathbf{dL}_{h,k}^c := \left[ \left. \frac{d\, L_h^c(\phi)}{d\, \phi} \right|_{\phi_h(\mathbf{f}(\mathbf{x}^j))} \cdot \left. \frac{\partial \phi_h(\mathbf{f})}{\partial f_k} \right|_{\mathbf{f}(\mathbf{x}^j)} \right]'_{\mathbf{x}^j \in \mathcal{S}},$$

the gradient of the penalty term is

$$\nabla_k V(\mathbf{f}) = \sum_{h=1}^{H} \frac{\lambda_h^v}{|\mathcal{S}|} \cdot \mathbf{K}_k \cdot \mathbf{dL}_{h,k}^c \ .$$

and, finally, we get

$$\nabla_k\, E(\mathbf{f}) = \mathbf{K}_k \cdot \left[ \frac{\lambda_k^\tau}{|\mathcal{L}_k|} \cdot \mathbf{I}_k^L \cdot \mathbf{dL}_k^e + +2 \cdot \lambda_k^r \cdot \mathbf{w}_k + \sum_{h=1}^{H} \frac{\lambda_h^v}{|\mathcal{S}|} \cdot \mathbf{dL}_{h,k}^c \right] \ . \qquad (3)$$

If $\mathbf{K}_k > 0$, the term in square brackets of equation (3) is null on any stationary point of $E(\cdot)$. This is a system of $k$ matrix equations, each involving $|\mathcal{S}|$ variables and scalar equations. The last term originating from the constraints correlates these equations. When optimizing via gradient descent, it is preferrable to drop the multiplication by $\mathbf{K}_k$ needed to obtain the exact gradient in order to avoid the stability issues that could be introduced by an ill-conditioned $\mathbf{K}_k$.

Whereas the use of a positive kernel would guarantee strict convexity when restricting the learning to the supervised examples as in standard kernel machines, $E(\cdot)$ is non-convex in any non trivial problem involving the constraint term. The labeled examples and the constraints are nominally coherent since they represent different reinforcing expressions of the concepts to be learned. Formally, $\forall \mathbf{x} \in \bigcup_k \mathcal{S}_k^L$, we have $\phi_h(f_1(\mathbf{x}), \ldots, f_T(\mathbf{x})) \geq 0$, which yields $L_h^c(f_1(\mathbf{x}), \ldots, f_T(\mathbf{x})) = 0$. As a result, the coherence condition suggests that the penalty term should be small when restricted to the supervised portion of the training set, after having learned the supervised examples. Hence, we propose to learn according to the following two stages:

1. Piagetian initialization : During this phase, we only enforce a regularized fitting of the supervised examples, by setting $\lambda_h^v = 0, h = 1, \ldots, H$, and $\lambda_k^\tau = \lambda^\tau, \lambda_k^r = \lambda^r, k = 1, \ldots, T$, where $\lambda^\tau$ and $\lambda^r$ are positive constants. This phase terminates according to standard stopping criteria adopted for plain kernel machines.
2. Abstraction : during this phase, the constraints are enforced in the cost function by setting $\lambda_h^v = \lambda^v, h = 1, \ldots, H$, where $\lambda^v$ is a positive constant. $\lambda^\tau$, $\lambda^r$ are not changed.

As explained in [5], this is related to some developmental psychology studies, which have shown that children experiment a stage-based learning. The two stages turn out to be a powerful way of tackling complexity issues and suggest a process in which the higher abstraction required to incorporate the constraints must follow the classic induction step that relies on supervised examples.

## 3 Logic constraints

In order to introduce logic clauses in the proposed learning framework, we can rely on the classic association from Boolean variables to real-valued functions by using the *t-norms* (triangular norms) [6]. A t-norm is any function $T : [0,1] \times [0,1] \to I\!R$, that is commutative (i.e. $T(x,y) = T(y,x)$), associative (i.e. $T(x, T(y,z)) = T(T(x,y),z)$), monotonic (i.e. $y \leq z \Rightarrow T(x,y) \leq T(x,z)$), and featuring a neutral element 1 (i.e. $T(x,1) = x$). A t-norm fuzzy logic is defined by its t-norm $T(x,y)$ that models the logic AND, while the negation of a variable $\neg x$ is computed as $1 - x$. The $t$-conorm, modeling the logical OR, is defined as $1 - T((1-x),(1-y))$, as a generalization of the De Morgan's law ($x \vee y = \neg(\neg x \wedge \neg y)$). Many different t-norm logics have been proposed in the literature. In the following we will consider the product t-norm $T(x,y) = x \cdot y$, but other choices are possible. In this case the t-conorm is computed as $1 - (1-x)(1-y) = x + y - xy$. Once the logic clauses are expressed using a t-norm, the constraint can be enforced by introducing a penalty that forces each clause to assume the value 1 on the given examples. Since t-norms are defined for input variables in $[0,1]$, whereas the functions $f_k(\mathbf{x})$ can take any real value, we apply a squashing function to constrain their values in $[0,1]$. Hence, the $h$-th logic clause can be enforced by the correspondent real-valued constraint,

$$t_h(\sigma(f_1(\mathbf{x})), \ldots, \sigma(f_T(\mathbf{x}))) - 1 \geq 0 \quad \forall \mathbf{x} \in \mathcal{S} \ , \qquad (4)$$

where $t_h(y_1, \ldots, y_T)$ is the implementation of the clause using the given t-norm and $\sigma : I\!R \to [0,1]$ is an increasing squashing function.

In order to have a more immediate compatibility with respect to the definition of t-norms, it is possible to exploit the targets $\{0,1\}$ for the $\{false, true\}$ values in the supervised examples. The use of these targets yields also an impact on the problem formulation. In fact, the regularization term tends to favor a constant solution equal to 0, that in this case biases the solution towards the $false$ value. This may be an useful property for those cases in which the negative class is not well described by the given examples, as it happens for instance in verification tasks (i.e. false positives have to be avoided as much as possible). In this case, a natural choice for the squash function is the piece linear mapping $\sigma(y) = \min(1, \max(y, 0))$. This is the setting we exploited in the experimental evaluation, but it is straightforward to redefine the task in an unbiased setting by mapping the logic values to $\{-1, 1\}$.

The constraints of equation (4) can be enforced during learning by using an appropriate loss function that penalizes their violation. In this case we can define

$$L_h^c(\phi_h(\mathbf{f}(\mathbf{x}))) = 1 - t_h(\sigma(f_1(\mathbf{x})), \ldots, \sigma(f_T(\mathbf{x}))) \ .$$

since the penalty is null only when the t-norm expression assumes exactly the value 1 and positive in the other cases.

When the available knowledge is represented by a set of propositions $C_1, \ldots, C_H$ that must jointly hold, we can enforce these constraints as separate penalties on their t-norm implementations or by combining the propositions in an unique constraint by considering the implementation of the only proposition $C = C_1 \wedge C_2 \wedge \ldots \wedge C_H$. The first choice is more flexible since it allows us to give different weights to each constraint and to realize different policies for activating the constraints during the learning process. This observation allows us to generalize the implementation to any logical constraint written in Conjunctive Normal Form (CNF). Let's consider a disjunction of a set of variables

$$\bigvee_{i \in P} a_i \vee \bigvee_{j \in N} \neg a_j = \neg \left( \bigwedge_{i \in P} \neg a_i \wedge \bigwedge_{j \in N} a_j \right) \ ,$$

where $P$ and $N$ are the sets of asserted and negated literals that appear in the proposition. If we implement the proposition using the product t-norm, we get

$$t_h(a_1, \ldots, a_T) = 1 - \prod_{i \in N_h} a_i \cdot \prod_{j \in P_h} (1 - a_j), \ h = 1, \ldots, H,$$

where $P_h$ and $N_h$ are the sets of asserted and negated literals. The conjunction of the single terms in a CNF can be directly implemented by multiplying the associated t-norm expressions $t_h(a_1, \ldots, a_T)$, but, as stated before, the minimization of $1 - C(a_1, \ldots, a_T)$ can be also performed by jointly minimizing the expressions $1 - t_h(a_1, \ldots, a_T)$, that force each term of the conjunction to be true. The derivative of each term can be computed easily as

$$\sigma'(f_k) \cdot \prod_{i \in N_h / \{k\}} \sigma(f_i) \cdot \prod_{j \in P_h} (1 - \sigma(f_j))$$

when $k \in N_h$, and

$$-\sigma'(f_k) \cdot \prod_{i \in N_h} \sigma(f_i) \cdot \prod_{j \in P_h/\{k\}} (1 - \sigma(f_j))$$

when $k \in P_h$, where $\sigma'(f_k)$ is the derivative of the squash function. Since all the factors in the products are non-negative, the previous derivatives are non-negative when $k \in N_h$, and non-positive when $k \in P_h$.

Finally, it is worth mentioning that each penalty by itself has a number of global minima related to the input configurations that make true the corresponding logic proposition. Hence, the resulting cost function is also likely to be plagued by the presence of multiple local minima, and, consequently, it is needed to devise ad-hoc optimization techniques to find good solutions for most of the problems.

## 4 Experimental results

This section presents a detailed experimental analysis on some artificial benchmarks properly created to stress the comparisons with plain kernel machines. All the generated datasets assume equiprobable classes and uniform density distributions over hyper-rectangles. Therefore, let $C$ be the number of classes and $N$ the total number of available examples, each class is represented by $\frac{N}{C}$ examples of which half positive and half negative. Furthermore, we assume to have available some prior knowledge on the classification task expressed by a set of logic clauses. The two-stage learning algorithm described in section 2 is exploited in all the experiments and, unless otherwise stated, all learned models are based on a Gaussian kernel with fixed $\sigma$ set to 0.4. This choice is motivated by the goal of comparing the proposed method with respect to plain kernel machines, rather than yielding the best performances. All benchmarks are based on a test set of 100 patterns per class, which are selected via the same sampling schema used to generate the corresponding training set. All presented results are an average over multiple runs performed on different instances of the training and test sets.

### 4.1 Benchmark 1: exponentially increasing class regions

This synthetic experiment aims at analyzing the effect of the presence of the a priori knowledge, implemented in the constraints, when the examples get sparser in the feature space. Let us assume to have $n$ classes, $C_1, \ldots, C_n$. The patterns for each class are uniformly sampled from a square in $I\!R^2$ centered in $(0, 0)$. Let $l > 0$ be the length of the side of the square for class $C_1$. The side of the square increases of a constant factor $\alpha > 1$ as we move from $C_i$ to $C_{i+1}$. Therefore, patterns of $C_i$ are sampled from a square of side length $l\alpha^i$, whose area grows moving up in the class order $i$ as $\alpha^{2i}$. Using a Gaussian kernel with fixed variance, this dataset would also require the number of labeled patterns for class $C_i$ to grow exponentially as we move to the higher order classes, to keep an adequate coverage in the feature space. This is required to model the higher variability of the input patterns that are distributed over a much larger area. However, labeled data is often scarce in real world applications and we model this fact by assuming that a fixed number of supervised examples is provided for each class. In this experiment, we study how the learner copes with the the patterns getting sparser, making the generalization more difficult.

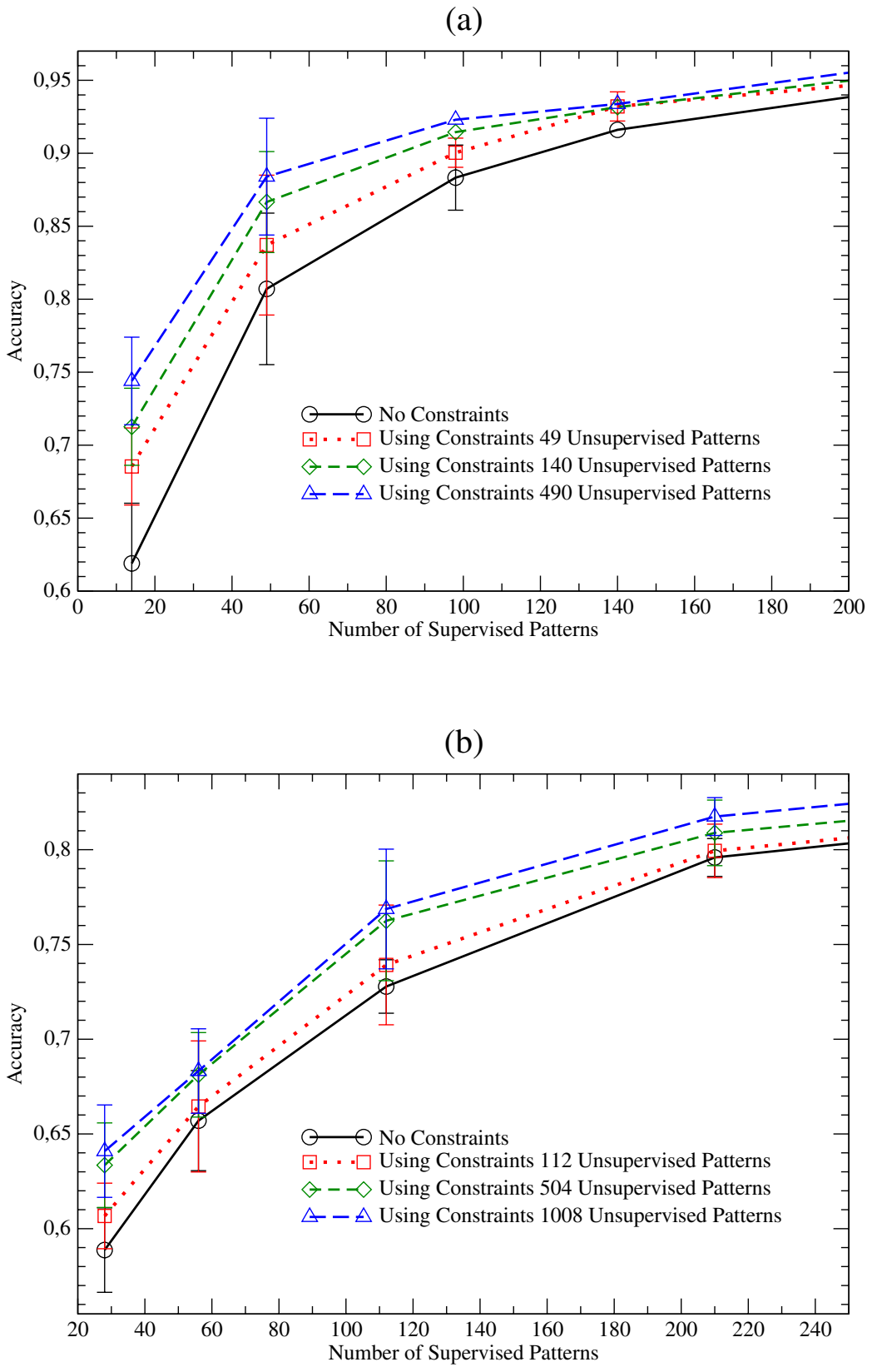

**Figure 1.** Benchmark 1: the accuracy values when using 7 (a) and 14 (b) classes.

To test the accuracy gain introduced by learning with constraints, we assume to have available some prior knowledge about the "inclusion" relationship of the regions of the classes: patterns of class $C_i$ cover an area that is included inside the area spanned by the patterns of class $C_{i+1}$. This knowledge can be expressed in form of logic constraints as: $i = 1, ..., n-1$, $\forall \mathbf{x}, c_i(\mathbf{x}) \Rightarrow c_{i+1}(\mathbf{x})$, where $c_i(\mathbf{x})$ is a unary predicate stating whether pattern $\mathbf{x}$ belongs to class $C_i$. For sake of compactness, we will refer to the i-th proposition as $c_i \Rightarrow c_{i+1}$. The same compact notation will be used to represent any logic clause also in the following of the paper. We also assume to know a-priori that any pattern must belong to at least one class (Closed World Assumption). This can be stated in logical form as: $\bigvee_{i=1}^{n} c_i$

We compared the classification accuracy against a standard kernel machine, which does not integrate the constraints directly during learning. However, the standard kernel machine exploits the a-priori knowledge via a simple pre-processing of the training pattern labels: if a pattern $x$ is a supervised example for the $i^{th}$ class $C_i$, then it is a supervised example also for each class $C_j$ with $j > i$. This is commonly done to process a hierarchy of classes (in our experiment the taxonomy reduces to a simple sequence).

Figure 1 plots the classification accuracy over the test set for $n = 7$ and $n = 14$, as average over 10 different instances of the supervised, unsupervised and test patterns. The growth parameter *alpha* has been set to 1.3 for this experiment. A t-student test confirms that the accuracy improvement for the learner, for which the logic constraints are enforced, is statistically significant for small labeled sets and when using a large number of unlabeled patterns, showing that the constraints are able to provide an effective aid for adequately covering the class regions when the supervised examples are scarce.

## 4.2 Benchmark 2: 3 classes, 2 clauses

This experiment aims at analyzing the effects on the classification accuracy due to the use of the logic constraints, when varying the dimension of the feature space. In particular, it consists of a multi-class classification task with 3 different classes $(A, B, C)$, which are known (a-priori) to be arranged according to a hierarchy defined by the clauses $a \wedge b \Rightarrow c$ and $a \vee b \vee c$. The patterns for each class lay in a hyper-rectangle in $\mathbb{R}^n$, where the dimensionality $n$ was varied in $\{3, 7, 10\}$. Given an uniform sampling over the hyper-rectangles, a higher dimensional input space corresponds to sparser training data for a fixed number of labeled patterns. This is an effect of the well known *curse-of-dimensionality*, making generalization more difficult in high dimensional input spaces.

In particular, the classes are defined according to the following geometry:

$$A = \{\mathbf{x} : 0 \le x_1 \le 2, 0 \le x_2 \le 2, 0 \le x_3 \le 1, \ldots, 0 \le x_n \le 1\}$$
$$B = \{\mathbf{x} : 1 \le x_1 \le 3, 0 \le x_2 \le 2, 0 \le x_3 \le 1, \ldots, 0 \le x_n \le 1\}$$
$$C = \{\mathbf{x} : 1 \le x_1 \le 2, 0 \le x_2 \le 2, 0 \le x_3 \le 1, \ldots, 0 \le x_n \le 1\}$$

During different runs of the experiment, the training set size has been increased from 6 to 480 examples and the unsupervised data ranged from 0 to 1350 patterns. In order to reduce the sampling noise, the accuracy values have been averaged over 6 different instances of the supervised, unsupervised and test sets.

Figure 2-(a) compares the classification accuracy obtained when the patterns lay in $\mathbb{R}^3$. The plot reports only the results for a maximum of 100 supervised patterns. Indeed, the learning task is trivially determined when abundant supervised data is available and there is little gain from enforcing constraints. This is consistent with the fact that the trained kernel machine is known to converge to the Bayes optimal classifier when the number of training examples tends to infinity. For sake of clearness, we also omitted the curve with 1350 unsupervised patterns as the gain over using 480 unsupervised is negligible. Figures 2-(b) and 2-(c) plot the classification accuracy obtained for patterns in $\mathbb{R}^7$ and $\mathbb{R}^{10}$, respectively. When moving to higher dimensional spaces, the learning task is harder and the accuracy gain grows to approximatively 20%. The gain would ultimately reduce when further increasing the training data, but this would have required a huge number of training patterns (which are rarely available in real-world applications).

## 4.3 Benchmark 3: 4 classes and 2 clauses

This multi-class classification task consists of 4 different classes: $A, B, C, D$. The patterns for each class are assumed to be uniformly distributed on a hyper-rectangle in $\mathbb{R}^n$, according to the following set definitions:

$$A = \{x : 0 \le x_1 \le 3, 0 \le x_2 \le 3, 0 \le x_3 \le 1, \ldots, 0 \le x_n \le 1\}$$
$$B = \{x : 1 \le x_1 \le 4, 1 \le x_2 \le 4, 0 \le x_3 \le 1, \ldots, 0 \le x_n \le 1\}$$
$$C = \{x : 2 \le x_1 \le 5, 2 \le x_2 \le 5, 0 \le x_3 \le 1, \ldots, 0 \le x_n \le 1\}$$
$$D = \{x : 1 \le x_1 \le 3, 1 \le x_2 \le 3, 0 \le x_3 \le 1, \ldots, 0 \le x_n \le 1 \vee 2 \le x_1 \le 4, 2 \le x_2 \le 4, 0 \le x_3 \le 1, \ldots, 0 \le x_n \le 1\}$$

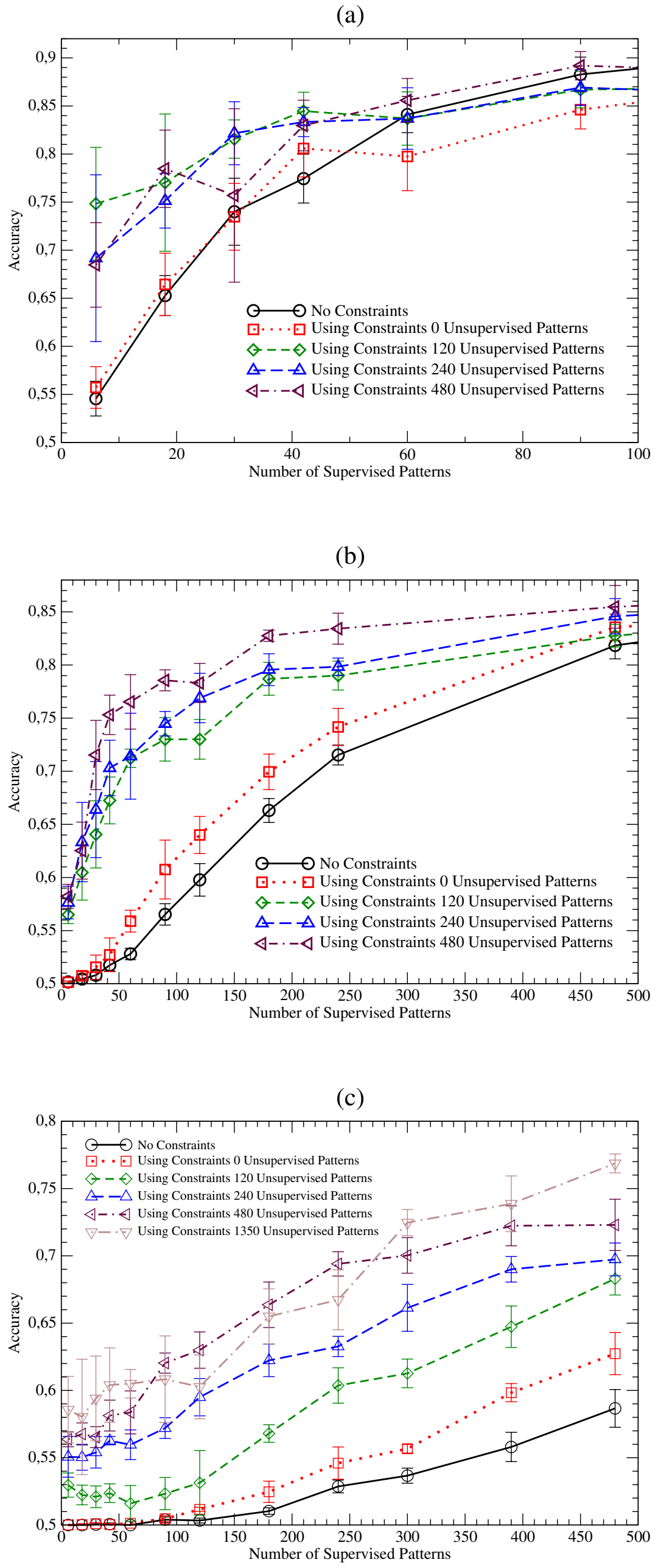

**Figure 2.** Benchmark 2: classification accuracy when using or not using the constraints varying the size of the labeled and unlabeled datasets for patterns laying in $\mathbb{R}^3$ (a), $\mathbb{R}^7$ (b) and $\mathbb{R}^{10}$ (c).

The following clauses are supposed to be known a-priori about the geometry of the classification task: $(a \wedge b) \vee (b \wedge c) \Rightarrow d$ and $a \vee b \vee c \vee d$. The first clause was converted in CNF and both constraints were directly integrated into the learning task as explained in section 3.

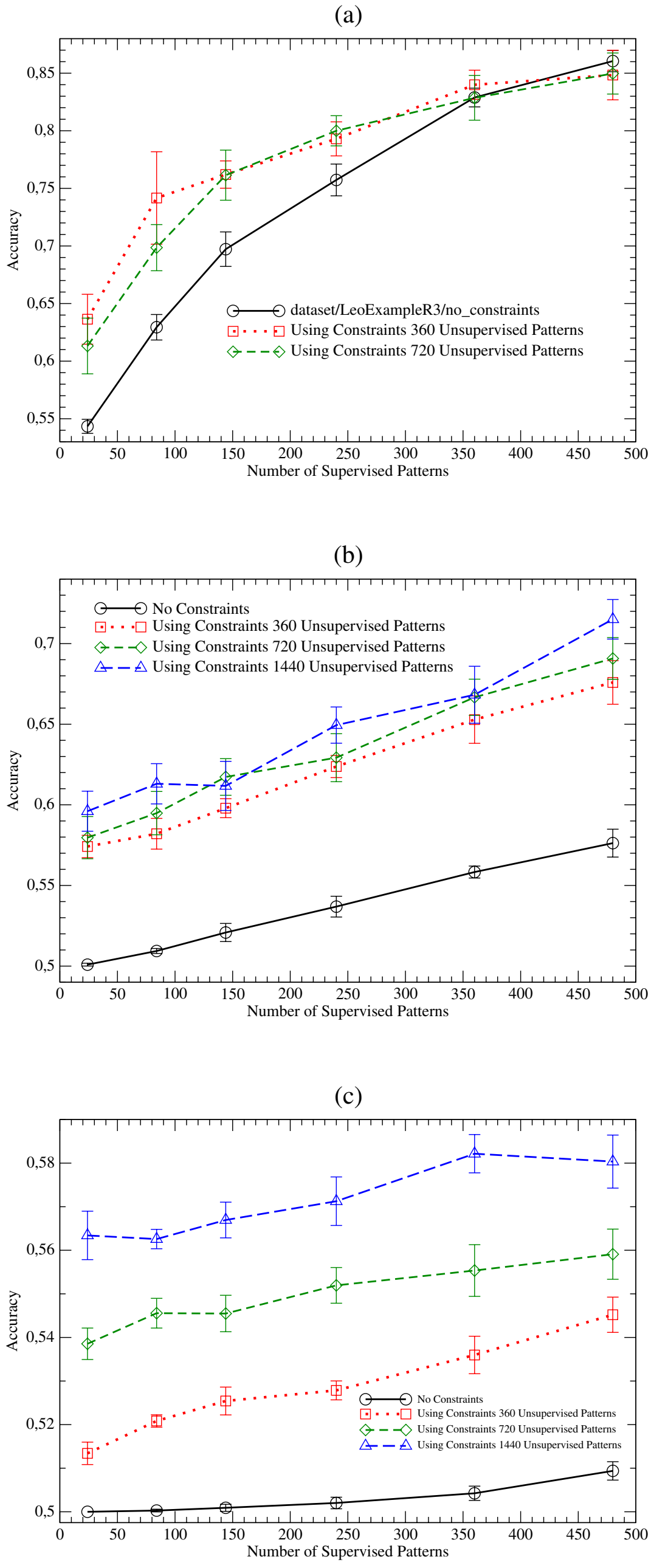

**Figure 3.** Benchmark 3: classification accuracy when using or not using the constraints varying the size of the labeled and unlabeled datasets for patterns laying in $I\!R^3$ (a),$I\!R^7$ (b) and $I\!R^{10}$ (c).

Figure 3 reports the classification accuracy in generalization, obtained when using the constraints and the unsupervised data versus the case when no constraints are employed in the learning procedure. In particular, the figures 3-(a), 3-(b) and 3-(c) report the classification accuracy (averaged over 6 random data generations), when the patterns are defined in $I\!R^3$, $I\!R^7$ and $I\!R^{14}$, respectively. The classifier trained using the constraints outperforms the one learned without the constraints by a statistically significant margin, which becomes very significant in higher dimensional spaces, where standard kernel machines based on a Gaussian kernel can not generalize without a very high number of labeled patterns.

## 5 Conclusions and future work

This paper presented a novel framework for bridging logic and kernel machines by extending the general apparatus of regularization with the introduction of logic constraints in the learning objective. If the constraint satisfaction is relaxed to be explicitly enforced only on the supervised and unsupervised examples, a representation theorem holds which dictates that the optimal solution of the problem is still a kernel expansion over the available examples. This allows the definition of a semi-supervised scheme in which the unsupervised examples help to approximate the penalty term associated with the logic constraints. While the optimization of the error functions deriving from the proposed formulation is plagued by local minima, we show successful results on artificial benchmarks thanks to a stage-based learning inspired to developmental psychology. This result reinforce the belief on the importance of the gradual presentation of examples [1]. The experimental analysis aims at studying the effect of the introduction of the constraints in the learning process for different dimensionalities of the input space, showing that the accuracy gain is very significant for larger input spaces, corresponding to harder learning settings, where generalization using standard kernel machines is often difficult. The proposed framework opens the doors to a new class of *semantic-based regularization machines* in which it is possible to integrate prior knowledge using high level abstract representations, including logic formalisms.

## REFERENCES

[1] Y. Bengio, 'Curriculum learning', in *Proceedings of the 26th Annual International Conference on Machine Learning*, pp. 41–48, (2009).

[2] A. Caponnetto, C.A. Micchelli, M. Pontil, and Y. Ying, 'Universal Kernels for Multi-Task Learning', *Journal of Machine Learning Research*, (2008).

[3] O. Chapelle, 'Training a support vector machine in the primal', *Neural Computation*, **19**(5), 1155–1178, (2007).

[4] P. Frasconi and A. Passerini, 'Learning with kernels and logical representations', in *Probabilistic Inductive Logic Programming: Theory and Applications, De Raedt, L. et al Eds*, ed., Springer, pp. 56–91, (2008).

[5] M. Gori, 'Semantic-based regularization and Piaget's cognitive stages', *Neural Networks*, **22**(7), 1035–1036, (2009).

[6] E.P. Klement, R. Mesiar, and E. Pap, *Triangular Norms*, Kluwer Academic Publisher, 2000.

[7] S. Muggleton, Lodhi H., A. Amini, and M.J.E. Sternberg, 'Support vector inductive logic programming', in *A. Hoffmann, H. Motoda, and T. Scheffer (Eds.):*, ed., Morgan Kaufmann, pp. 163–175, (2005).

[8] B. Scholkopf and A. J. Smola, *Learning with Kernels*, MIT Press, Cambridge, MA, USA, 2001.